\pdfoutput=1

\documentclass[11pt]{article}

\usepackage[]{EMNLP2022}

\usepackage{times}
\usepackage{latexsym}

\usepackage{tabularx}
\usepackage{booktabs}
\usepackage{graphicx}
\usepackage{multirow}
\usepackage{tablefootnote}
\usepackage{subcaption}
\usepackage[utf8]{inputenc}

\usepackage{microtype}
\usepackage{hyperref}

\usepackage{amsmath}
\usepackage{amssymb}
\DeclareMathOperator*{\argmin}{argmin}

\usepackage{inconsolata}
\usepackage{array}
\newcolumntype{H}{>{\setbox0=\hbox\bgroup}c<{\egroup}@{}}
\usepackage{bm}
\usepackage{subfiles}

\usepackage[T1,T2A]{fontenc}
\usepackage[english,russian,ukrainian]{babel}
\usepackage{kotex}

\title{Multilingual Relation Classification via Efficient and Effective Prompting}

\author{Yuxuan Chen ~~~ David Harbecke ~~~ Leonhard Hennig\\
German Research Center for Artificial Intelligence (DFKI)  \\
Speech and Language Technology Lab\\
\{\textit{yuxuan.chen, david.harbecke, leonhard.hennig}\}@\textit{dfki.de}}

\begin{document}
\selectlanguage{english}
\newcommand{\insertfullysupervised}{
    \begin{table*}[h!]
        \setlength{\tabcolsep}{4.8pt}
		\scriptsize
		\centering
		\begin{tabular}{@{}lHH rrrrrrrrrrrrrr|rrrr|c@{}}
			\toprule
			Method & PLM & \#Params & AR & DE & EN & ES & FA & FR & IT & KO & NL & PL & PT & RU & SV & UK & EN & H & M & L & $\overline X$\\
			\midrule
			EN(B) & & &-&-&94.9&-&-&-&-&-&-&-&-&-&-&-& 94.9 &-&-&-&-\\
			
			\textsc{XLM-R}\textsubscript{\textsc{EM}} & XLM-R\textsubscript{\textsc{Base}} & 125M & \textbf{98.4}&\textbf{95.7}&95.9&27.9&0.0&82.6&\textbf{98.9}&64.6&92.2&\textbf{97.4}&\textbf{97.4}&\textbf{96.9}&2.2&5.1&95.9 & 86.1 & 54.3 & 3.7&68.2\\
			
			null prompts & \multirow{4}*{mT5\textsubscript{\textsc{Base}}} & 125M &85.5&81.6&84.7&59.8&71.2&82.6&84.2&63.3&71.4&49.4&12.9&84.9&48.9&46.2&84.7 & 65.8 & 73.3 & 47.6&66.2\\
			\midrule
			
			CS & & \multirow{3}*{220M} &95.1&95.4&96.0&74.7&69.2&\textbf{97.2}&98.3&82.1&\textbf{96.9}&94.8&95.3&87.6&48.9&46.2&96.0 & \textbf{92.5} & 82.1 & 47.6&84.1\\
			
			SP & &
			&95.1&88.5&\textbf{96.1}&\textbf{81.1}&65.4&97.0&97.1&83.1&59.9&95.6&96.9&87.3&\textbf{63.0}&51.3&\textbf{96.1} & 87.9 & 81.2 & 57.2 &82.7\\
			
			IL & & &94.1&94.0& 96.0 &70.5&\textbf{73.1}&97.2&97.0&\textbf{83.2}&93.5&93.0&85.2&83.3&58.7&\textbf{71.8}&96.0 & 89.2 & \textbf{83.5} & \textbf{65.2}&\textbf{85.0}\\\bottomrule

		\end{tabular}
		\caption{Fully-supervised results in micro-F1 (\%) on the SMiLER dataset. The evaluated methods are the proposed baseline EN(B)~\cite{seganti_2021_smiler}, \textsc{XLM-R}\textsubscript{\textsc{EM}}, null prompts, and ours. EN, H, M, L: macro average across the languages within the respective group.
		$\overline X$: macro average across all 14 languages.
		Our variants outperform all baselines along all groups averages, \textsc{XLM-R}\textsubscript{\textsc{EM}} has good results for many high-resource languages.
		Overall, in-language prompting performs best, especially for lower-resource languages.}
		\label{tab:fully-supervised}
	\end{table*}
}

\newcommand{\insertfewshotgroup}{
    \begin{table}[t!]
		\footnotesize
		\centering
		\begin{tabular}{@{}cl rrrr r@{}}
			\toprule
			Shots & Method & EN & H & M & L & $\overline X$\\
			\midrule
			 
			 \multirow{5}*{8} & \textsc{XLM-R}\textsubscript{\textsc{EM}}  &31.8 & \textbf{43.0} & 27.5 & 6.6& 33.7\\
			
			 & null prompts  & 37.4 & 27.6 & 26.6 & 37.4&29.5 \\\cmidrule{2-7}
			
			 & CS  & 42.2 & 30.6 & 27.8 & 38.4& 32.0\\
			 & SP  & \textbf{45.4} & 27.8 & 17.9 & 33.6&27.4 \\
			 & IL & 42.2 & 40.5 & \textbf{38.3} & \textbf{43.4}& \textbf{40.6}\\\midrule
			 
			 \multirow{5}*{16} & \textsc{XLM-R}\textsubscript{\textsc{EM}}  & \textbf{56.4} & \textbf{56.9} & 34.1 & 10.4 &45.3 \\
			
			 & null prompts  & 42.1 & 31.6 & 34.3 & 49.7 & 35.5\\\cmidrule{2-7}
			
			 & CS  & 50.5 & 50.1 & 41.9 & 53.9 & \textbf{48.9}\\
			 & SP  & 53.7 & 46.7 & 38.4 & 49.0 & 45.8\\
			 & IL & 50.5 & 45.2 & \textbf{42.1} & \textbf{54.6} & 46.3\\
			 \midrule
			 
			 \multirow{5}*{32} & \textsc{XLM-R}\textsubscript{\textsc{EM}}  & 73.2 & 62.4 & 44.4 & 6.5 & 51.3\\
			
			 & null prompts  & 56.0 & 36.4 & 47.7 & 53.9 & 42.7\\\cmidrule{2-7}
			
			 & CS  & \textbf{80.9} & 57.0 & \textbf{65.1} & 59.4 & 60.8\\
			 & SP  & 61.2 & 53.5 & 46.3 & 63.1 & 53.9\\
			 & IL & \textbf{80.9} & \textbf{63.6} & 64.2 & \textbf{67.4} & \textbf{65.5}\\
			 \bottomrule

		\end{tabular}
		\caption{Few-shot results by group in micro-F1 (\%) on the SMiLER~\cite{seganti_2021_smiler} dataset averaged over five runs.
		We macro-average results for each language group (see Figure \ref{fig:lang-dist}) and over all languages ($\overline X$).
		In-language prompting performs best in most settings and language groups. Our variants are especially strong for medium- and lower-resource language groups.
		See Table~\ref{tab:few-shot-variance} in Appendix~\ref{sec:additional_tables} for detailed results with mean and std.\ for each language.}
		\label{tab:group}
	\end{table}
}

\newcommand{\insertfewshot}{
    \begin{table*}[h!]
    \resizebox{\columnwidth}{!}{
		\centering
		\begin{tabular}{@{}cl rrrrrrrrrrrrrr|c@{}}
			\toprule
			Shots & Method & AR & DE & EN & ES & FA & FR & IT & KO & NL & PL & PT & RU & SV & UK & $\overline X$\\
			\midrule
			
			\multirow{5}*{8} &
			\textsc{XLM-R}\textsubscript{\textsc{EM}} &\textbf{58.8} & \textbf{49.2} & 31.8 & 12.8 & 7.3 & 30.6 & \textbf{52.3} &16.5 & 38.1 & \textbf{46.6} & \textbf{53.8} & \textbf{60.7} & 1.3 & 12.0 & 33.7\\
			\cmidrule{2-17}
			&null prompts & 17.2 & 28.1 & 37.4 & 10.4 & 25.8 & 14.6 & 28.0 & \textbf{36.9} & 44.4 & 29.7 & 26.1 & 39.2 & 47.4 & 27.4 & 29.5\\
			\cmidrule{2-17}
			&CS & 19.6 & 11.1 & 42.2 & 26.2 & 45.0 & 36.3 & 42.3 & 18.7 & 28.0 & 26.7 & 27.0 & 47.5 & \textbf{48.9} & 28.0 & 32.0\\
			& SP & 14.2 & 29.1 & \textbf{45.4} & 32.5 & 18.8 & 20.1 & 26.8 &20.6 & 31.8 & 26.3 & 26.0 & 29.6 & 36.4 & 30.8 & 27.4\\
			& IL &33.4 & 39.0 & 42.2 & \textbf{37.9} & \textbf{46.0} & \textbf{39.1} & 35.1 &35.5 & \textbf{52.5} & 31.8 & 32.8 & 55.9 & 34.1 & \textbf{52.8} & \textbf{40.6}\\
			\midrule
			
			\multirow{5}*{16} &
			\textsc{XLM-R}\textsubscript{\textsc{EM}} &\textbf{67.7} & 44.3 & \textbf{56.4} & 19.6 & 7.8 & 47.5 & \textbf{76.1} & 26.7 & 64.7 & \textbf{62.8} & \textbf{69.1} & \textbf{70.9} & 1.3 & 19.5 & 45.3\\\cmidrule{2-17}
			&null prompts & 34.5 & 18.1 & 42.1 & 20.5 & 43.2 & 28.7 & 38.0 & 25.3 & 37.5 & 37.8 & 17.8 & 54.3 & \textbf{56.6} & 42.8 & 35.5\\\cmidrule{2-17}
			& CS & 36.6 & \textbf{62.5} & 50.5 & 26.1 & \textbf{49.7} & 47.3 & 53.5 &\textbf{39.3} & \textbf{71.2} & 33.5 & 45.3 & 61.2 & 49.4 & \textbf{58.4} & \textbf{48.9}\\
			& SP & 38.6 & 40.2 & 53.7 & \textbf{52.0} & 37.9 & \textbf{51.3} & 46.6 & 38.6 & 40.2 & 53.7 & 52.0 & 37.9 & 51.3 & 46.6 & 45.8\\
			& IL &47.0 & 62.5 & 50.5 & 31.1 & 45.6 & 21.7 & 32.8 & 33.7 & 39.2 & 58.5 & 50.2 & 65.4 & 51.1 & 58.2 & 46.3\\
			\midrule
			
			\multirow{5}*{32} &
			\textsc{XLM-R}\textsubscript{\textsc{EM}} & \textbf{81.6} & 59.9 & 73.2 & 21.4 & 12.7 & 58.8 & \textbf{81.0}&38.8 & \textbf{74.5} & \textbf{77.7} & 63.2 & 62.5 & 1.3 & 11.7 & 51.3\\\cmidrule{2-17}
			&null prompts & 45.4 & 26.0 & 56.0 & 14.3 & 67.4 & 48.6 & 42.8 & 30.2 & 54.9 & 40.7 & 15.1 & 48.4 & 49.7 & \textbf{58.1} & 42.7\\\cmidrule{2-17}
			& CS & 62.0 & \textbf{72.1} & \textbf{80.9} & 40.6 & 61.0 & 51.4 & 50.4 &\textbf{72.2} & 71.4 & 39.0 & \textbf{73.3} & 57.7 & 67.6 & 51.3 & 60.8\\
			& SP & 50.3 & 35.5 & 61.2 & \textbf{60.8} & 59.0 & \textbf{74.2} & 34.7 &29.6 & 42.7 & 67.4 & 47.4 & \textbf{65.1} & \textbf{69.2} & 57.0 & 53.9\\
			& IL & 65.5 & 61.9 & 80.9 & 53.1 & \textbf{76.4} & 62.0 & 71.7 & 50.8 & 71.3 & 65.2 & 59.5 & 63.8 & 63.6 & 71.1 & \textbf{65.5}\\
			\bottomrule
		\end{tabular}
		}
		\caption{Few-shot results in micro-F1 (\%) on the SMiLER dataset. We evaluate \textsc{XLM-R}\textsubscript{\textsc{EM}}, null prompts, and our variants. \textsc{XLM-R}\textsubscript{\textsc{EM}} is \textsc{BERT}\textsubscript{\textsc{EM}}~\cite{baldini-soares_2019_matching} but with \textsc{XLM-R}\textsubscript{\textsc{Base}} as encoder. Null prompts and ours use mT5\textsubscript{\textsc{Base}}. For each result, the mean and standard deviation of 5 runs are reported. $\overline X$: macro average across 14 languages. "CS": code-switch prompting. "SP": insert soft tokens into code-switch prompts. "IL": in-language prompting.}
		\label{tab:few-shot}
	\end{table*}
}

\newcommand{\insertfewshotvariance}{
    \begin{table}[h]
    \onecolumn
    \centering
    \resizebox{0.97\linewidth}{!}{
		\begin{tabular}{@{}cl ccccccc|c@{}}
			\toprule
			Shots & Method & AR & DE & EN & ES & FA & FR & IT &\\
			\cmidrule{1-9}
			
			\multirow{5}*{8} &
			\textsc{XLM-R}\textsubscript{\textsc{EM}} &\textbf{58.8}\scriptsize{$\pm$20.2} & \textbf{49.2}\scriptsize{$\pm$7.2} & 31.8\scriptsize{$\pm$11.3} & 12.8\scriptsize{$\pm$6.4} & 7.3\scriptsize{$\pm$4.6} & 30.6\scriptsize{$\pm$4.0} & \textbf{52.3}\scriptsize{$\pm$5.0}\\
			&null prompts & 17.2\scriptsize{$\pm$10.6} & 28.1\scriptsize{$\pm$16.6} & 37.4\scriptsize{$\pm$10.1} & 10.4\scriptsize{$\pm$7.9} & 25.8\scriptsize{$\pm$10.1} & 14.6\scriptsize{$\pm$10.4} & 28.0\scriptsize{$\pm$22.3}\\
			\cmidrule{2-9}
			&CS & 19.6\scriptsize{$\pm$10.2} & 11.1\scriptsize{$\pm$17.2} & 42.2\scriptsize{$\pm$17.5} & 26.2\scriptsize{$\pm$21.5} & 45.0\scriptsize{$\pm$12.0} & 36.3\scriptsize{$\pm$17.4} & 42.3\scriptsize{$\pm$4.9}\\
			& SP & 14.2\scriptsize{$\pm$5.5} & 29.1\scriptsize{$\pm$18.6} & \textbf{45.4}\scriptsize{$\pm$7.9} & 32.5\scriptsize{$\pm$12.1} & 18.8\scriptsize{$\pm$12.0} & 20.1\scriptsize{$\pm$11.0} & 26.8\scriptsize{$\pm$19.1}\\
			& IL &33.4\scriptsize{$\pm$25.4} & 39.0\scriptsize{$\pm$19.3} & 42.2\scriptsize{$\pm$17.5} & \textbf{37.9}\scriptsize{$\pm$15.3} & \textbf{46.0}\scriptsize{$\pm$28.5} & \textbf{39.1}\scriptsize{$\pm$15.9} & 35.1\scriptsize{$\pm$19.0}\\
			\cmidrule{1-9}
			
			\multirow{5}*{16} &
			\textsc{XLM-R}\textsubscript{\textsc{EM}} &\textbf{67.7}\scriptsize{$\pm$17.5} & 44.3\scriptsize{$\pm$23.1} & \textbf{56.4}\scriptsize{$\pm$4.2} & 19.6\scriptsize{$\pm$6.5} & 7.8\scriptsize{$\pm$9.4} & 47.5\scriptsize{$\pm$8.4} & \textbf{76.1}\scriptsize{$\pm$4.3}\\
			&null prompts & 34.5\scriptsize{$\pm$18.4} & 18.1\scriptsize{$\pm$20.4} & 42.1\scriptsize{$\pm$15.5} & 20.5\scriptsize{$\pm$12.0} & 43.2\scriptsize{$\pm$14.9} & 28.7\scriptsize{$\pm$22.0} & 38.0\scriptsize{$\pm$18.9}\\\cmidrule{2-9}
			& CS & 36.6\scriptsize{$\pm$18.1} & \textbf{62.5}\scriptsize{$\pm$11.0} & 50.5\scriptsize{$\pm$32.3} & 26.1\scriptsize{$\pm$21.5} & \textbf{49.7}\scriptsize{$\pm$11.1} & 47.3\scriptsize{$\pm$30.3} & 53.5\scriptsize{$\pm$27.7}\\
			& SP & 38.6\scriptsize{$\pm$17.6} & 40.2\scriptsize{$\pm$29.4} & 53.7\scriptsize{$\pm$25.2} & \textbf{52.0}\scriptsize{$\pm$13.8} & 37.9\scriptsize{$\pm$14.0} & \textbf{51.3}\scriptsize{$\pm$27.3} & 46.6\scriptsize{$\pm$24.3}\\
			& IL &47.0\scriptsize{$\pm$32.3} & 62.5\scriptsize{$\pm$11.0} & 50.5\scriptsize{$\pm$32,3} & 31.1\scriptsize{$\pm$22.2} & 45.6\scriptsize{$\pm$23.1} & 21.7\scriptsize{$\pm$17.5} & 32.8\scriptsize{$\pm$18.1}\\
			\cmidrule{1-9}
			
			\multirow{5}*{32} &
			\textsc{XLM-R}\textsubscript{\textsc{EM}} & \textbf{81.6}\scriptsize{$\pm$9.4} & 59.9\scriptsize{$\pm$29.8} & 73.2\scriptsize{$\pm$4.4} & 21.4\scriptsize{$\pm$3.1} & 12.7\scriptsize{$\pm$6.3} & 58.8\scriptsize{$\pm$10.1} & \textbf{81.0}\scriptsize{$\pm$2.6}\\
			&null prompts & 45.4\scriptsize{$\pm$20.0} & 26.0\scriptsize{$\pm$24.3} & 56.0\scriptsize{$\pm$13.4} & 14.3\scriptsize{$\pm$15.1} & 67.4\scriptsize{$\pm$6.3} & 48.6\scriptsize{$\pm$16.8} & 42.8\scriptsize{$\pm$21.0}\\\cmidrule{2-9}
			& CS & 62.0\scriptsize{$\pm$26.7} & \textbf{72.1}\scriptsize{$\pm$15.0} & \textbf{80.9}\scriptsize{$\pm$4.3} & 40.6\scriptsize{$\pm$30.3} & 61.0\scriptsize{$\pm$28.2} & 51.4\scriptsize{$\pm$22.1} & 50.4\scriptsize{$\pm$37.9}\\
			& SP & 50.3\scriptsize{$\pm$19.7} & 35.5\scriptsize{$\pm$33.8} & 61.2\scriptsize{$\pm$29.3} & \textbf{60.8}\scriptsize{$\pm$26.7} & 59.0\scriptsize{$\pm$27.1} & \textbf{74.2}\scriptsize{$\pm$12.5} & 34.7\scriptsize{$\pm$36.2}\\
			& IL & 65.5\scriptsize{$\pm$22.0} & 61.9\scriptsize{$\pm$29.5} & 80.9\scriptsize{$\pm$4.3} & 53.1\scriptsize{$\pm$28.5} & \textbf{76.4}\scriptsize{$\pm$9.9} & 62.0\scriptsize{$\pm$29.1} & 71.7\scriptsize{$\pm$26.6}\\
			\midrule
			
            \toprule
			 &  & KO & NL & PL & PT & RU & SV & UK & $\overline X$\\
			\midrule
			
			\multirow{5}*{8} &
			\textsc{XLM-R}\textsubscript{\textsc{EM}} &16.5\scriptsize{$\pm$7.9} & 38.1\scriptsize{$\pm$10.3} & \textbf{46.6}\scriptsize{$\pm$10.0} & \textbf{53.8}\scriptsize{$\pm$3.8} & \textbf{60.7}\scriptsize{$\pm$7.2} & 1.3\scriptsize{$\pm$0.8} & 12.0\scriptsize{$\pm$7.0} & 33.7\\
			&null prompts & \textbf{36.9}\scriptsize{$\pm$14.1} & 44.4\scriptsize{$\pm$8.5} & 29.7\scriptsize{$\pm$15.2} & 26.1\scriptsize{$\pm$17.0} & 39.2\scriptsize{$\pm$12.3} & 47.4\scriptsize{$\pm$14.5} & 27.4\scriptsize{$\pm$15.6} & 29.5\\\cmidrule{2-10}
			& CS & 18.7\scriptsize{$\pm$17.4} & 28.0\scriptsize{$\pm$15.6} & 26.7\scriptsize{$\pm$16.6} & 27.0\scriptsize{$\pm$14.1} & 47.5\scriptsize{$\pm$15.4} & \textbf{48.9}\scriptsize{$\pm$16.6} & 28.0\scriptsize{$\pm$17.5} & 32.0\\
			& SP &20.6\scriptsize{$\pm$19.4} & 31.8\scriptsize{$\pm$14.7} & 26.3\scriptsize{$\pm$16.2} & 26.0\scriptsize{$\pm$15.2} & 29.6\scriptsize{$\pm$16.6} & 36.4\scriptsize{$\pm$25.3} & 30.8\scriptsize{$\pm$28.8} & 27.4\\
			& IL &35.5\scriptsize{$\pm$19.9} & \textbf{52.5}\scriptsize{$\pm$4.2} & 31.8\scriptsize{$\pm$12,9} & 32.8\scriptsize{$\pm$19.3} & 55.9\scriptsize{$\pm$14.7} & 34.1\scriptsize{$\pm$26.9} & \textbf{52.8}\scriptsize{$\pm$14.2} & \textbf{40.6}\\
			\midrule
			
			\multirow{5}*{16} &
			\textsc{XLM-R}\textsubscript{\textsc{EM}} &26.7\scriptsize{$\pm$5.0} & 64.7\scriptsize{$\pm$2.8} & \textbf{62.8}\scriptsize{$\pm$5.6} & \textbf{69.1}\scriptsize{$\pm$2.8} & \textbf{70.9}\scriptsize{$\pm$8.6} & 1.3\scriptsize{$\pm$0.4} & 19.5\scriptsize{$\pm$12.2} & 45.3\\
			&null prompts & 25.3\scriptsize{$\pm$19.0} & 37.5\scriptsize{$\pm$14.3} & 37.8\scriptsize{$\pm$8.8} & 17.8\scriptsize{$\pm$16.1} & 54.3\scriptsize{$\pm$20.3} & \textbf{56.6}\scriptsize{$\pm$23.5} & 42.8\scriptsize{$\pm$20.6} & 35.5\\\cmidrule{2-10}
			& CS &\textbf{39.3}\scriptsize{$\pm$9.4} & \textbf{71.2}\scriptsize{$\pm$9.0} & 33.5\scriptsize{$\pm$25.0} & 45.3\scriptsize{$\pm$19.1} & 61.2\scriptsize{$\pm$26.1} & 49.4\scriptsize{$\pm$24.7} & \textbf{58.4}\scriptsize{$\pm$22.4} & \textbf{48.9}\\
			& SP & 38.6\scriptsize{$\pm$17.6} & 40.2\scriptsize{$\pm$29.4} & 53.7\scriptsize{$\pm$25.2} & 52.0\scriptsize{$\pm$13.8} & 37.9\scriptsize{$\pm$14.0} & 51.3\scriptsize{$\pm$27.3} & 46.6\scriptsize{$\pm$24.3} & 45.8\\
			& IL &33.7\scriptsize{$\pm$20.2} & 39.2\scriptsize{$\pm$11.4} & 58.5\scriptsize{$\pm$18.9} & 50.2\scriptsize{$\pm$19.4} & 65.4\scriptsize{$\pm$6.5} & 51.1\scriptsize{$\pm$22.9} & 58.2\scriptsize{$\pm$20.9} & 46.3\\
			\midrule
			
			\multirow{5}*{32} &
			\textsc{XLM-R}\textsubscript{\textsc{EM}} &38.8\scriptsize{$\pm$3.3} & \textbf{74.5}\scriptsize{$\pm$2.8} & \textbf{77.7}\scriptsize{$\pm$1.6} & 63.2\scriptsize{$\pm$26.2} & 62.5\scriptsize{$\pm$12.8} & 1.3\scriptsize{$\pm$1.3} & 11.7\scriptsize{$\pm$5.9} & 51.3\\
			&null prompts & 30.2\scriptsize{$\pm$29.8} & 54.9\scriptsize{$\pm$25.6} & 40.7\scriptsize{$\pm$21.7} & 15.1\scriptsize{$\pm$16.9} & 48.4\scriptsize{$\pm$33.7} & 49.7\scriptsize{$\pm$30.7} & \textbf{58.1}\scriptsize{$\pm$27.4} & 42.7 \\\cmidrule{2-10}
			& CS &\textbf{72.2}\scriptsize{$\pm$6.9} & 71.4\scriptsize{$\pm$23.5} & 39.0\scriptsize{$\pm$30.0} & \textbf{73.3}\scriptsize{$\pm$8.0} & 57.7\scriptsize{$\pm$20.3} & 67.6\scriptsize{$\pm$12.0} & 51.3\scriptsize{$\pm$23.7} & 60.8\\
			& SP &29.6\scriptsize{$\pm$34.7} & 42.7\scriptsize{$\pm$33.5} & 67.4\scriptsize{$\pm$12.3} & 47.4\scriptsize{$\pm$28.0} & \textbf{65.1}\scriptsize{$\pm$19.0} & \textbf{69.2}\scriptsize{$\pm$20.1} & 57.0\scriptsize{$\pm$32.1} & 53.9\\
			& IL & 50.8\scriptsize{$\pm$24.9} & 71.3\scriptsize{$\pm$12.5} & 65.2\scriptsize{$\pm$25.8} & 59.5\scriptsize{$\pm$28.5} & 63.8\scriptsize{$\pm$27.4} & 63.6\scriptsize{$\pm$26.1} & 71.1\scriptsize{$\pm$17.4} & \textbf{65.5}\\
			\bottomrule
		\end{tabular}
		}
		\caption{Few-shot results in micro-F1 (\%) on the SMiLER dataset. We evaluate \textsc{XLM-R}\textsubscript{\textsc{EM}}, null prompts, and our prompt variants.
		For each result, the mean and standard deviation of 5 runs are reported.
		$\overline X$: macro average across 14 languages.
		The standard deviations are quite large which indicates that multiple runs are needed and results are seed dependent.
		In-language prompting provides the most consistent results, with Polish 8-shot as lowest score (31.8 $F_1$).
		Other methods all have results below 15.0 $F_1$.}
	\label{tab:few-shot-variance}
	\end{table}	
}

\newcommand{\insertzeroshot}{
\begin{table*}[t!]
		\footnotesize
		\centering
		\begin{tabular}{@{}cl rrrrrrrrrrrrrr@{}}
			\toprule
			 & & EN & AR & DE & ES & FA & FR & IT & KO & NL & PL & PT & RU & SV & UK\\
			\midrule
			
			\multicolumn{2}{l}{Random}&2.8&11.1&4.6&4.8&12.5&4.6&4.6&3.6&4.6&4.8&4.6&12.5&4.6&14.3\\\midrule
			
			\multicolumn{16}{c}{\textit{Zero-Shot In-Context Learning}}\\
			\midrule
			
			\multirow{2}*{SVO} 
			& CS &\multirow{2}{*}{\textbf{5.5}}&\textbf{69.9}&\textbf{10.4}&12.7&38.5&\textbf{13.3}&11.2&\textbf{10.0}&\textbf{12.4}&\textbf{14.0}&8.1&52.3&27.2&\textbf{51.3}\\
			& IL &&2.2&5.2&1.8&5.3&9.2&1.3&3.6&7.6&9.0&1.7&7.1&5.4&25.6\\
			\midrule
			
			\multirow{2}*{SOV} &
			CS &\multirow{2}{*}{4.8}&68.4&10.0&\textbf{13.2}&36.9&12.3&\textbf{12.6}&5.0&11.8&13.4&\textbf{10.3}&\textbf{52.6}&\textbf{29.4}&\textbf{51.3}\\
			& IL &&3.8&5.0&3.6&\textbf{59.8}&7.7&1.3&3.1&10.0&7.9&1.4&6.0&4.5&25.6\\
			\midrule
			
			\multicolumn{16}{c}{\textit{Zero-Shot Cross-Lingual Transfer}}\\
			\midrule
			
			\multicolumn{2}{c}{EN (268K)} & - & 94.0 & 94.9 & 91.7 & 91.1 & 96.0 & 97.5 & 78.2 & 97.5 & 93.3 & 95.2 & 93.8 & 97.8 & 94.7\\
			\multicolumn{2}{c}{EN-small (36K)} & - & 45.9 & 64.7 & 73.1 & 70.3 & 82.2 & 77.5 & 30.8 & 79.9 & 59.0 & 67.3 & 76.1 & 77.2 & 54.1\\
			\bottomrule
			
		\end{tabular}
		\caption{Zero-shot results in micro-F1 (\%) on the SMiLER dataset. "SVO" and "SOV": word order of prompting. Overall, Code-switch prompting performs the best in the zero-shot in-context scenario. In cross-lingual transfer experiments, English-task training greatly improves the performance on all the other 13 languages.}
		\label{tab:zero-shot}
	\end{table*}
}

\newcommand{\insertsmiler}{
\begin{table}[t]
    \resizebox{\columnwidth}{!}{
    \setlength{\tabcolsep}{10pt}
	\centering
	\begin{tabular}{@{}lrrHrrr@{}}
	\toprule
	\multirow{2}*{Lang.} & \multicolumn{4}{l}{Fine-tuning data} & \multicolumn{2}{l}{Pre-train tokens}\\\cmidrule{2-7}
	&\multicolumn{1}{l}{\#Class} & \multicolumn{1}{l}{\#Train(K)} & \#Test & Max. & mT5(B) & XLM-R(B)\\
	\midrule
	AR & 9 & 9.3 & 190 & 74 & 57 & 2.9\\
	DE & 22 & 51.5 & 1051 & 84 & 347 & 10.3\\
	EN & 36 & 267.6 & 5461 & 110 & 2733 & 55.6\\
	ES & 21 & 11.1 & 226 & 70 & 433 & 9.4\\
	FA & 8 & 2.6 & 54 & 93 & 52 & 13.3\\
	FR & 22 & 60.9 & 1243 & 83 & 318 & 9.8\\
	IT & 22 & 74.0 & 1510 & 86 & 162 & 5.0\\
	KO & 28 & 18.7 & 382 & 95 & 26 & 5.6\\
	NL & 22 & 38.9 & 793 & 76 & 73 & 5.0\\
	PL & 21 & 16.8 & 344 & 86 & 130 & 6.5\\
	PT & 22 & 43.3 & 885 & 82 & 146 & 8.4\\
	RU & 8 & 6.4 & 131 & 69 & 713 & 23.4\\
	SV & 22 & 4.5 & 92 & 84 & 45 & 0.08\\
	UK & 7 & 1.0 & 20 & 65 & 41 & 0.006\\
    \bottomrule
	\end{tabular}
	}
	\caption{Statistics of the 14 languages in the SMiLER dataset, including the number of classes, the number of training examples (in thousands), and the maximum text length over train and test splits. Appended to the table are the sizes (in billion tokens) of pre-training corpora of the referred languages for mT5 and XLM-R, respectively.}
	\label{tab:smiler}
\end{table}
}

\newcommand{\insertverbalizations}{
\begin{table*}[ht]
	\scriptsize
	\centering
	\begin{tabular}{@{}llll rr@{}}
	\toprule
	\multirow{2}*{Task} &
	\multirow{2}*{Dataset} & \multirow{2}*{\#Class} & \multirow{2}*{Verbalizations} & \multicolumn{2}{c}{\# Token in Verb.}\\
	\cmidrule{5-6}
	 & & & & Mean & Std.\\
	\midrule
	LA & CoLA~\cite{warstadt_2019_cola} & 2 & \texttt{correct, incorrect.}~\cite{gao_2021_making} & 1 & 0\\
	NER & CoNLL03~\cite{sang_2003_conll} & 5 & \texttt{location, person, not an, ...}~\cite{cui_2021_template} & 1.2 & 0.4 \\
	NLI & MNLI~\cite{williams_2018_mnli} & 3 & \texttt{yes, no, maybe.}~\cite{fu_2022_polyglot} & 1 & 0 \\
	NLI & XNLI~\cite{conneau_2018_xnli} & 3 & \texttt{yes, no, maybe; Evet, ...}~\cite{zhao_2021_discrete} & 1 & 0 \\
	PI & PAWS-X~\cite{yang_2019_paws} & 2 & \texttt{yes, no.}~\cite{qi_2022_enhancing} & 1 & 0\\
	TC & MARC~\cite{keung_2020_marc} & 2 & \texttt{good, \{average, bad\}.}~\cite{huang_2022_zero} & 1 & 0\\
	
	\midrule
	\multirow{6}*{RC} &
	TACRED~\cite{zhang_2017_tacred} & 42 & \texttt{founded by, city of birth, country of death, ...} & 3.23 & 1.99\\
	 & SemEval~\cite{hendrickx_2010_semeval} & 10 & \texttt{cause effect, entity origin, product producer, ...} & 2.50 & 0.81\\
	 & NYT~\cite{riedel_2010_modeling} & 24 & \texttt{ethnicity, major shareholder of, religion, ...} & 2.10 & 1.01\\
	 & \textsc{SciERC}~\cite{luan_2018_scierc} & 6 & \texttt{conjuction, feature of, part of, used for, ...} & 2.17 & 0.69\\
	\cmidrule{2-6}
	 & SMiLER (EN)~\cite{seganti_2021_smiler} & 36 & \texttt{birth place, starring, won award, ...} & 2.58 & 0.68\\
	 & SMiLER (ALL)~\cite{seganti_2021_smiler} & 36 & \texttt{hat Genre, chef d'organisation, del país, ...} & 3.66 & 1.44\\
    \bottomrule
	\end{tabular}
	\caption{Statistics of the lengths of the verbalizations over several classification tasks. The lengths for non-RC tasks depend on the tokenizers from the respective PLMs in the cited work. The lengths for RC tasks are based on the mT5\textsubscript{\textsc{Base}} tokenizer. Mean and std. show that the label space of the RC task is more complex than most few-class classification tasks. The verbalizations of RC datasets are listed in Appendix~\ref{sec:verbalizers}. For SemEval, the two possible directions of a relation are combined. 
	For NYT, we use the version from \citet{zeng_2018_extracting}. For SMiLER, "EN" is the English split; "ALL" contains all data from 14 languages. }
	\label{tab:verbalizations}
\end{table*}
}

\newcommand{\insertprompts}{
\begin{table*}[t!]
    \resizebox{\textwidth}{!}{
	\footnotesize
	\centering
	\begin{tabular}{@{}lllll@{}}
	\toprule
	 & \multirow{2}*{Prompt input} & \multirow{2}*{Target} & \multicolumn{2}{c}{Example}\\
	 \cmidrule{4-5}
	& & &Input & Target\\
	\midrule
	null prompts & $\bm x. \,\,\_\_\_\_$ & $\phi^{EN}(r)$ & \textit{Goethe} schrieb \textit{Faust}. \_\_\_\_ & has author\\\midrule
 
	CS & $\bm x.\,\bm e_h \,\,\_\_\_\_\,\, \bm e_t$ & $\phi^{EN}(r)$ & \textit{Goethe} schrieb \textit{Faust}. \textit{Faust} \_\_\_\_ \textit{Goethe} & has author\\
	 
    SP & $\bm x.\,\texttt{[v\textsubscript{1}]}\bm e_h \,\,\texttt{[v\textsubscript{2}]}\_\_\_\_\,\, \texttt{[v\textsubscript{3}]}\bm e_t$ & $\phi^{EN}(r)$ & \textit{Goethe} schrieb \textit{Faust}. \texttt{[v\textsubscript{1}]}\textit{Faust} \texttt{[v\textsubscript{2}]}\_\_\_\_ \texttt{[v\textsubscript{3}]}\textit{Goethe} & has author\\
	
    IL & $\bm x.\,\bm e_h \,\,\_\_\_\_\,\, \bm e_t$ & $\phi^{\mathcal L}(r)$ & \textit{Goethe} schrieb \textit{Faust}. \textit{Faust} \_\_\_\_ \textit{Goethe} & hat Autor\\
    \bottomrule
	\end{tabular}
	}
	\caption{Overview of the prompts, including null prompts (baseline), and ours with its variants. For each prompt or its variant, we list (1) the prompt input and the target; (2) an example based on the plain text in German ``\textit{Goethe} schrieb \textit{Faust}.'' \texttt{[v\textsubscript{i}]}: learnable soft tokens. $\bm \phi^{EN}(r)$: the original (English) relation verbalization. $\phi^{\mathcal L}(r)$: the translated relation verbalization into the target language $\mathcal L$.
	}
	\label{tab:prompts}
\end{table*}
}
\maketitle

\begin{abstract}
Prompting pre-trained language models has achieved impressive performance on various NLP tasks, especially in low data regimes. Despite the success of prompting in monolingual settings, applying prompt-based methods in multilingual scenarios has been limited to a narrow set of tasks, due to the high cost of handcrafting multilingual prompts. In this paper, we present the first work on prompt-based multilingual relation classification (RC), by introducing an efficient and effective method that constructs prompts from relation triples and involves only minimal translation for the class labels. We evaluate its performance in fully supervised, few-shot and zero-shot scenarios, and analyze its effectiveness across 14 languages, prompt variants, and English-task training in cross-lingual settings. 
We find that in both fully supervised and few-shot scenarios, our prompt method beats competitive baselines: fine-tuning \textsc{XLM-R}\textsubscript{\textsc{EM}} and null prompts. It also outperforms the random baseline by a large margin in zero-shot experiments. Our method requires little in-language knowledge and can be used as a strong baseline for similar multilingual classification tasks.
\end{abstract}

\section{Introduction}
\label{sec:introduction}

Relation classification (RC) is a crucial task in information extraction (IE), aiming to identify the relation between entities in a text~\cite{alt_2019_fine}.  
Extending RC to multilingual settings has recently received increased interest~\cite{zou_2018_adversarial, kolluru_2022_alignment}, but the majority of prior work still focuses on English~\cite{baldini-soares_2019_matching,lyu-chen_2021_relation}.
A main bottleneck for multilingual RC is the lack of supervised resources, comparable in size to large English datasets \cite{riedel_2010_modeling, zhang_2017_tacred}.
The SMiLER dataset \cite{seganti_2021_smiler} provides a starting point to test fully supervised and more efficient approaches due to different resource availability for different languages. 

Previous studies have shown the promising performance of prompting PLMs compared to the data-hungry fine-tuning, especially in low-resource scenarios~\cite{gao_2021_making, le-scao-rush_2021_many, lu_2022_fantastically}. Multilingual pre-trained language models~\cite{conneau_2020_unsupervised, xue_2021_mt5} further enable multiple languages to be represented in a shared semantic space, thus making prompting in multilingual scenarios feasible. However, the study of prompting for multilingual tasks so far remains limited to a small range of tasks such as text classification~\cite{winata_2021_language} and natural language inference~\cite{lin_2022_few}.
To our knowledge, the effectiveness of prompt-based methods for multilingual RC is still unexplored.

To analyse this gap, we pose two research questions for multilingual RC with prompts:\\
\textbf{RQ1.} What is the most effective way to prompt? We investigate whether prompting should be done in English or the target language and whether to use soft prompt tokens.\\
\textbf{RQ2.} How well do prompts perform in different data regimes and languages? 
We investigate the effectiveness of our prompting approach in three scenarios: fully supervised, few-shot and zero-shot.
We explore to what extent the results are related to the available language resources.

\begin{figure*}[t]
\centering
\includegraphics[width=0.98\textwidth,keepaspectratio]{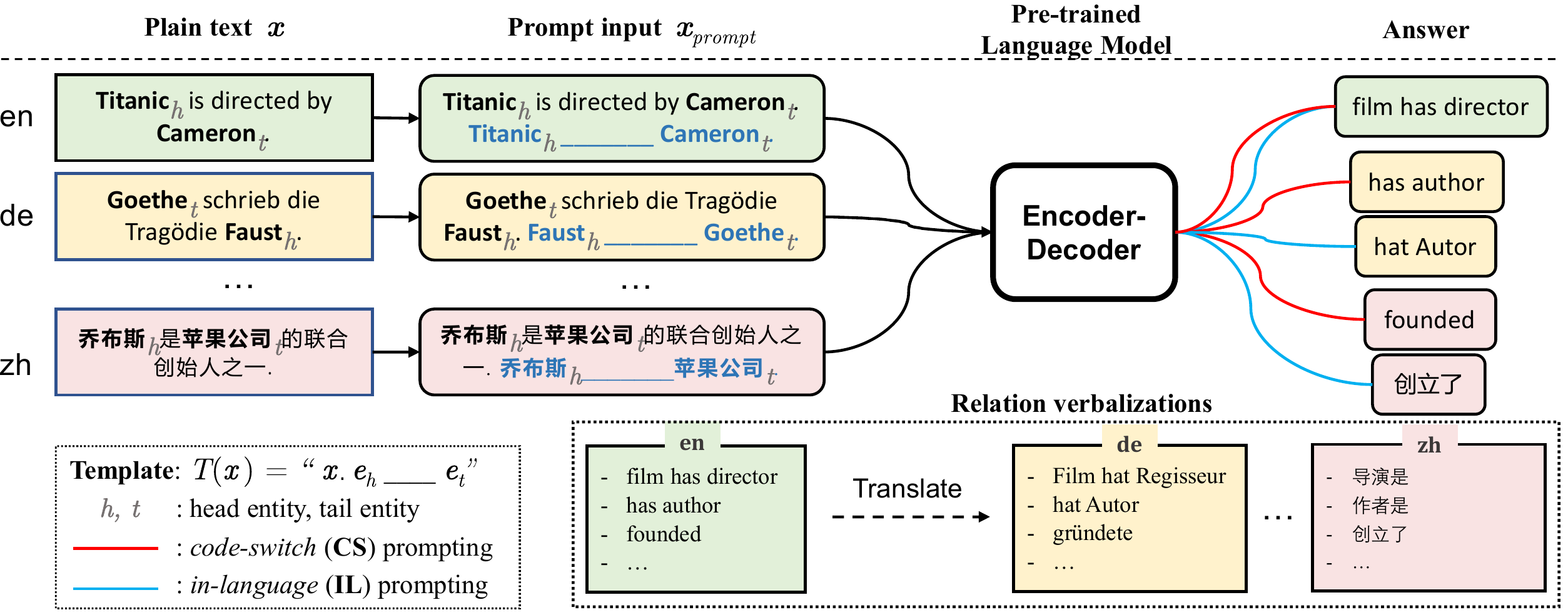}
\caption{Overview of our approach. Given a plain text $\bm x$ containing head entity $\bm e_h$ and tail entity $\bm e_t$ from language $\mathcal L$, we first apply the template $T(\bm x) = ``\bm x.\,\bm e_h \,\,\_\_\_\_\,\, \bm e_t$'' and yield the prompt input with a blank. Then the PLM aims to fill in the relation at the blank. In code-switch prompting, the target sequence is the English relation verbalization. In in-language prompting, the target is the relation name translated into $\mathcal L$.}
\label{fig:overview}
\end{figure*}

We present an efficient and effective prompt method for multilingual RC (see Figure \ref{fig:overview}) that derives prompts from relation triplets (see Section~\ref{sec:template}).
The derived prompts include the original sentence and entities and are supposed to be filled with the relation label.
We evaluate the prompts with three variants, two of which require no translation, and one of which requires minimal translation, i.e., of the relation labels only.
We find that our method outperforms fine-tuning and a strong task-agnostic prompt baseline in fully supervised and few-shot scenarios, especially for relatively low-resource languages. Our method also improves over the random baseline in zero-shot settings, and achieves promising cross-lingual performance. The main contributions of this work hence are:
\begin{itemize}
  \itemsep0em
  \item We propose a simple but efficient prompt method for multilingual RC, which is, to the best of our knowledge, the first work to apply prompt-based methods to multilingual RC  (Section~\ref{sec:methods}).
  \item We evaluate our method on the largest multilingual RC dataset, SMiLER ~\cite{seganti_2021_smiler}, and compare our method with strong baselines in all three scenarios. We also investigate the effects of different prompt variants, including insertion of soft tokens, prompt language, and the word order of prompting (Sections~\ref{sec:experiments} \&~\ref{sec:results}).
\end{itemize}

\section{Preliminaries}
\label{sec:preliminaries}

We first give a formal definition of the relation classification task, and then introduce fine-tuning and prompting paradigms to perform RC.

\subsection{Relation Classification Task Definition}
\label{sec:task-formulation}
Relation classification is the task of classifying the relationship such as \textit{date\_of\_birth}, \textit{founded\_by} or \textit{parents} between pairs of entities in a given context.
	
Formally, given a relation set $\mathcal R$ and a text $\bm x=[x_1, x_2, \ldots, x_{n}]$ (where $x_1, \cdots, x_{n}$ are tokens) with two disjoint spans $\bm e_h$ and $\bm e_t$ denoting the head and tail entity, RC aims to predict the relation $r\in\mathcal R$ between $\bm e_h$ and $\bm e_t$, or give a \textit{no\_relation} prediction if no relation in $\mathcal R$ holds. RC is a multilingual task if the token sequences come from different languages.

\subsection{Fine-tuning for Relation Classification}
In fine-tuning, a task-specific linear classifier is added on top of the PLM.
Fine-tuning hence introduces a different scenario from pre-training, since language model (LM) pre-training is usually formalized as a cloze-style task to predict target tokens at \texttt{[MASK]}~\cite{devlin_2019_bert,liu_2019_roberta} or a corrupted span~\cite{raffel_2020_t5,lewis_2020_bart}.
For the RC task, the classifier aims to predict the target class $r$ at \texttt{[CLS]} or at the entity spans denoted by \textsc{Marker} ~\cite{baldini-soares_2019_matching}.

\subsection{Prompting for Relation Classification}

\insertprompts

Prompting is proposed to bridge the gap between pre-training and fine-tuning~\cite{liu_2021_pre-train, gu_2022_ppt}. The essence of prompting is, by appending extra text to the original text according to a task-specific template $T(\cdot)$, to reformulate the downstream task to an LM pre-training task such as masked language modeling (MLM), and apply the same training objective during the task-specific training. 
For the RC task, to identify the relation between ``Angela Merkel'' and ``Joachim Sauer'' in the text ``Angela Merkel's current husband is quantum chemist Joachim Sauer,'' an intuitive template for prompting can be ``The relation between Angela Merkel and Joachim Sauer is \texttt{[MASK]},'' and the LM is supposed to assign a higher likelihood to the term \emph{couple} than to e.g.\ \emph{friends} or \emph{colleagues} at \texttt{[MASK]}. This ``fill-in the blank'' paradigm is well aligned with the pre-training scenario, and enables prompting to better coax the PLMs for pre-trained knowledge~\cite{petroni_2019_language}.

\section{Methods}
\label{sec:methods}

We now present our method, as shown in Figure~\ref{fig:overview}. We introduce its template and verbalizer, and propose several variants of the prompt. Lastly, we explain the training and inference process.

\subsection{Template}
\label{sec:template}
For prompting~\cite{liu_2021_pre-train}, a prompt often consists of a template $T(\cdot)$ and a verbalizer $\mathcal V$. Given a plain text $\bm x$, the template $T$ adds task-related instruction to $\bm x$ to yield the prompt input
\begin{equation}
   \bm x_{prompt}=T(\bm x). 
\end{equation}

Following \citet{chen_2021_knowledge} and \citet{han_2021_ptr}, we treat relations as predicates and use the cloze ``$\bm e_h$ \underline{ \{\emph{relation}\} } $\bm e_t$'' for the LM to fill in. Our template is formulated as 
\begin{equation}
T(\bm x) := \text{``}\bm x.\,\bm e_h \,\,\_\_\_\_\,\, \bm e_t\text{''}.
\end{equation}
In the template $T(\bm x)$, $\bm x$ is the original text and the two entities $\bm e_h$ and $\bm e_t$ come from $\bm x$. Therefore, our template does not introduce extra tokens, thus involves no translation at all.

\subsection{Verbalizer}
After being prompted by $\bm x_{prompt}$, the PLM $\mathcal M$ predicts the masked text $\bm y$ at the blank. To complete an NLP classification task, a verbalizer $\phi$ is required to bridge the set of labels $\mathcal Y$ and the set of predicted texts (verbalizations $\mathcal V)$.
For the simplicity of our prompt, we use the one-to-one verbalizer: 
\begin{equation}
\phi: \mathcal Y\to\mathcal V, r\mapsto \phi(r),
\end{equation}
where $r$ is a relation, and $\bm\phi(r)$ is the simple verbalization of $r$. $\phi(\cdot)$ normally only involves splitting $r$ by ``-'' or ``\_'' and replacing abbreviations such as \textit{org} with \textit{organization}. E.g., the relation \textit{org-has-member} corresponds to the verbalization ``organization has member''. 
Then the prediction is formalized as 
\begin{equation}
    p(r|\bm x) \propto p(\bm y=\phi(r)|\bm x_{prompt}; \theta_{\mathcal M}),
\end{equation}
where $\theta_{\mathcal M}$ denotes the parameters of model $\mathcal M$.
$p(r|\bm x)$ is normalized by the likelihood sum over all relations.

\subsection{Variants}

\insertverbalizations

To find the optimal way to prompt, we investigate three variants as follows.

\textbf{Hard prompt vs soft prompt (SP)}~~~Hard prompts (a.k.a.\ discrete prompts)~\cite{liu_2021_pre-train} are entirely formulated in natural language. Soft prompts (a.k.a.\ continuous prompts) consist of learnable tokens~\cite{lester_2021_power} that are not contained in the PLM vocabulary. Following \citet{han_2021_ptr}, we insert soft tokens before entities and blanks as shown for SP in Table~\ref{tab:prompts}.

\textbf{Code-switch (CS) vs in-language (IL)}~~~Relation labels are in English across almost all RC datasets. Given a text from a non-English input $\mathcal L$ with a blank, the recovered text is code-mixed after being completed with an English verbalization, corresponding to code-switch prompting. It is probably more reasonable for the PLM to fill in the blank in language $\mathcal L$. Inspired by \citet{lin_2022_few} and \citet{zhao_2021_discrete}, we machine-translate the English verbalizers into the other languages.\footnote{See Appendix~\ref{sec:verbalizers} for more examples of translated verbalizations. To translate the verbalizer of the SMiLER dataset, we use DeepL by default and Google Translate when the target language is not supported by DeepL (in case of AR, FA, KO and UK).} Table~\ref{tab:prompts} visualizes both code-switch (CS) and in-language (IL) prompting. For English, CS- and IL- prompting are equivalent, since $\mathcal L$ is English itself.

\textbf{Word order of prompting}~~~For the RC task, head-relation-tail triples involve three elements. Therefore, deriving natural language prompts from them requires handling where to put the predicate (relation). In the case of SOV languages, filling in a relation that occurs between $\bm e_h$ and $\bm e_t$ seems less intuitive. Therefore, to investigate if the word order of prompting affects prediction accuracy, we swap the entities and the blank in the SVO-template ``$\bm x.\,\bm e_h \,\,\_\_\_\_\,\, \bm e_t$'' and get ``$\bm x.\,\bm e_h \,\,\bm e_t \,\,\_\_\_\_$'' as the SOV-template.

\subsection{Training and Inference}
The training and inference setups depend on the employed model.
Prompting autoencoding language models requires the verbalizations to be of fixed length, since the length of masks, which is identical with verbalization length, is unknown during inference. Encoder-decoders can handle verbalizations of varying length by nature~\cite{han-2022-genpt,du_2021_all}.
\citet{han_2021_ptr} adjust all the verbalizations in TACRED to a length of 3, to enable prompting with RoBERTa for RC. We argue that for multilingual RC, this fix is largely infeasible, because: (1) in case of in-language prompting on SMiLER, the variance of the length of the verbalizations increases from 0.68 to 1.44 after translation (see Table~\ref{tab:verbalizations}), and surpasses most of listed monolingual RC datasets (SemEval, NYT and \textsc{SciERC}), making it harder to unify the length; (2) manually adjusting the translated prompts requires manual effort per target language, making it much more expensive than adjusting only English verbalizations. Therefore, we suggest using an encoder-decoder PLM for prompting~\cite{song_2022_clip}.

\textbf{Training objective}~~~For an encoder-decoder PLM $\mathcal M$, given the prompt input $T(\bm x)$ and the target sequence $\phi(r)$ (i.e. label verbalization), we denote the output sequence as $y$. The probability of an exact-match decoding is calculated as follows:
\begin{equation}
   \prod_{t=1}^{|\phi(r)|} P_{\theta}\left(y_t=\phi_t(r)|y_{<t}, T(\bm x)\right),
\end{equation}
where $y_t$, $\phi_t(r)$ denote the $t$-th token of $y$ and $\phi(r)$, respectively. $y_{<t}$ denotes the decoded sequence on the left. $\theta$ represents the set of all the learnable parameters, including those of the PLM $\theta_{\mathcal M}$, and those of the soft tokens $\theta_{sp}$ in case of variant ``soft prompt''.
Hence, the final objective over the training set $\mathcal X$ is to minimize the negative log-likelihood:\begin{equation}
\begin{split}
   \argmin_{\theta}&\, -\frac{1}{|\mathcal X|}\sum_{\bm x\in\mathcal X}\sum_{t=1}^{|\phi(r)|}\\
   &\log P_{\theta}\left(y_t = \phi_t(r)|y_{<t}, T(\bm x)\right).
\end{split}
\end{equation}

\textbf{Inference}~~~We collect the output logits of the decoder, $\mathsf L\in\mathbb R^{|V|\times L}$, where $|V|$ is the vocabulary size of $\mathcal M$, and $L$ is the maximum decode length. For each relation $r\in\mathcal R$, its score is given by~\cite{han-2022-genpt}:
\begin{equation}
    \mathrm{score}_{\theta}(r) := \frac{1}{|\phi(r)|}\sum_{t=1}^{|\phi(r)|} P_{\theta}(y_t=\phi_t(r)),
\end{equation}
where we compute $P$ by looking up in the $t$-th column of $\mathsf L$ and applying softmax at each time step $t$. We aggregate $P$ by addition to encourage partial matches as well, instead of enforcing exact matches. The score is normalized by the length of verbalization in order to avoid predictions favoring longer relations. Finally, we select the relation with the highest score as prediction.

\section{Experiments}
\label{sec:experiments}

We implement our experiments using the Hugging Face Transformers  library~\cite{wolf_2020_transformers}, Hydra~\cite{yadan_2019_hydra} and PyTorch~\cite{paszke_2019_pytorch}.\footnote{We make our code publicly available at \url{https://github.com/DFKI-NLP/meffi-prompt} for better reproducibility.}
We use micro-F1 as the evaluation metric, as the SMiLER paper~\cite{seganti_2021_smiler} suggests. To measure the overall performance over multiple languages, we report the macro average across languages, following \citet{zhao_2021_discrete} and \citet{lin_2022_few}.
We also group the languages by their available resources in both pre-training and fine-tuning datasets for additional aggregate results.
Details of the dataset, the models, and the experimental setups are as follows. Further experimental details are listed in Appendix~\ref{sec:experimental-details}.

\insertsmiler

\subsection{Dataset}
\label{sec:dataset}
We conduct an experimental evaluation of our multilingual prompt methods on the SMiLER ~\cite{seganti_2021_smiler} dataset, which contains 1.1M annotated texts across 14 languages.
\footnote{Note that SMiLER contains 3 versions of the English split: \textit{en} (268K training examples), \textit{en-small} (36K) and \textit{en-full} (744K). We use the \textit{en} version by default, unless specified otherwise.} Table~\ref{tab:smiler} lists the main statistics of the different languages in the SMiLER dataset. Note that languages have varying number of relations, mostly related to how many samples are present.
We do not evaluate other datasets because the only prior multilingual RC dataset that fits our task, RELX~\cite{koksal-ozgur_2020_relx}, contains only 502 parallel examples in 5 languages.

\begin{figure}[t]
 \centering
 \includegraphics[width=\columnwidth]{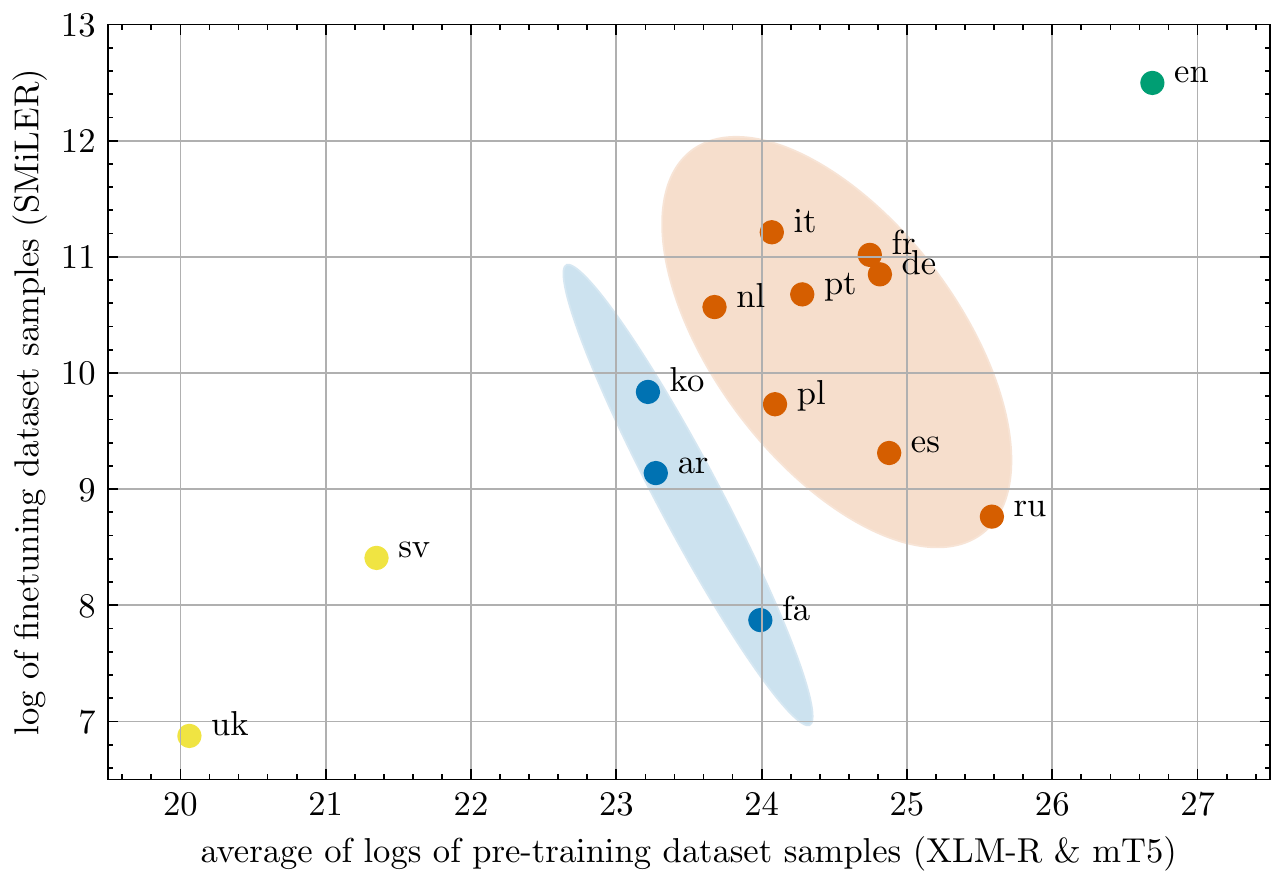}
 \caption{Pre-training and fine-tuning dataset size by language.
 Four languages groups are distinguishable:
 English (green) has by far the largest dataset, many other European languages (orange) have large datasets for pre-training and fine-tuning.
 The three non-European languages (blue) have either less pre-training or fine-tuning data and lowest resource are Swedish and Ukrainian (yellow).
 }
 \label{fig:lang-dist}
\end{figure}

\textbf{Grouping of the languages}~~~We visualize the languages in Figure~\ref{fig:lang-dist} based on the sizes of RC training data, but include the pre-training data as well, to give a more comprehensive overview of the availability of resources for each language.
We divide the 14 languages into 4 groups, according to the detectable clusters in Figure~\ref{fig:lang-dist} and language origins.

\subsection{Model}
\label{sec:model}

For prompting, we use mT5\textsubscript{\textsc{Base}}~\cite{xue_2021_mt5}, an encoder-decoder PLM that supports 101 languages, including all languages in SMiLER. mT5\textsubscript{\textsc{Base}}~\cite{xue_2021_mt5} has 220M parameters.

\subsection{Baselines}
\label{sec:baselines}
\hspace{\parindent}\textbf{EN(B)}~\cite{seganti_2021_smiler}~~~EN(B) is the baseline proposed together with the SMiLER dataset. They fine-tune BERT\textsubscript{\textsc{Base}} on the English training split and report the micro-F1 on the English test split. BERT\textsubscript{\textsc{Base}} has 110M parameters.

\textbf{\textsc{XLM-R}\textsubscript{\textsc{EM}}}~~~To provide a fine-tuning baseline, we re-implement \textsc{BERT}\textsubscript{\textsc{EM}}~\cite{baldini-soares_2019_matching} with the \textsc{Entity Start} variant.\footnote{We also open-source our implementation of \textsc{XLM-R}\textsubscript{\textsc{EM}} at \url{https://github.com/DFKI-NLP/mtb-bert-em}.} In this method, the top-layer representations at the starts of the two entities are concatenated for linear classification. To adapt \textsc{BERT}\textsubscript{\textsc{EM}} to multilingual tasks, we change the PLM from BERT to a multilingual autoencoder, XLM-R\textsubscript{\textsc{Base}}~\cite{conneau_2020_unsupervised}, and refer to this model as \textsc{XLM-R}\textsubscript{\textsc{EM}}. XLM-R\textsubscript{\textsc{Base}} has 125M parameters.

\textbf{Null prompts}~\cite{logan_2021_cutting}~~~To better verify the effectiveness of our method, we implement null prompts as a strong task-agnostic prompt baseline. Null prompts involve minimal prompt engineering by directly asking the LM about the relation, without giving any task instruction (see Table~\ref{tab:prompts}). \citet{logan_2021_cutting} show that null prompts surprisingly achieve on-par performance with handcrafted prompts on many tasks. For best comparability, we use the same PLM mT5\textsubscript{\textsc{Base}}.

\subsection{Fully Supervised Setup}
\label{sec:fully-supervised-setup}

We evaluate the performance of \textsc{XLM-R}\textsubscript{\textsc{EM}}, null prompts, and our method on each of the 14 languages, after training on the full train split from that language. The prompt input and target of null prompts and our prompts are listed in Table~\ref{tab:prompts}.

We employ the randomly generated seed 319 for all the evaluated methods. For \textsc{XLM-R}\textsubscript{\textsc{EM}}, we follow \citet{baldini-soares_2019_matching} and set the batch size to be 64, the optimizer to be Adam with the learning rate $3\times 10^{-5}$ and the number of epochs to be 5. For null prompts and ours, we use AdamW as the optimizer with the learning rate $3\times 10^{-5}$, as \citet{zhang_2022_downstream} suggest for most of the sequence-to-sequence tasks, the number of epochs to 5, and batch size to 16. The maximum sequence length is 256 for all methods.

\subsection{Few-shot Setup}
\label{sec:few-shot-setup}

Few-shot learning is normally cast as a $K$-shot problem, where $K$ labelled examples per class are available. We follow \citet{chen_2021_knowledge} and \citet{han_2021_ptr}, and evaluate on 8, 16 and 32 shots.

The few-shot training set $\mathcal D_{train}$ is generated by randomly sampling $K$ instances per relation from the training split.
The test set $\mathcal D_{test}$ is the original test split from that language. We follow \citet{gao_2021_making} and sample another $K$-shot set from the English train split as validation set $\mathcal D_{val}$. We tune hyperparameters on $\mathcal D_{val}$ for the English task, and apply these to all languages.

We evaluate the same methods as in the fully supervised scenarios, but repeat 5 runs as suggested in \citet{gao_2021_making}, and report the mean and standard deviation of micro-F1.
We use a fixed set of random seeds \{13, 36, 121, 223, 319\} for data generation and training across the 5 runs. 
For \textsc{XLM-R}\textsubscript{\textsc{EM}}, we use the same hyperparameters as \citet{baldini-soares_2019_matching}, a batch size of 256, and a learning rate of $1\times 10^{-4}$. For null prompts and our prompts, we set the learning rate to $3\times 10^{-4}$, batch size to 16, and the number of epochs to 20.

\subsection{Zero-shot Setup}
\label{sec:zero-shot-setup}
We consider two scenarios for zero-shot multilingual relation classification.

\textbf{Zero-shot in-context learning}~~~Following \citet{kojima_2022_large}, we investigate whether PLMs are also decent zero-shot reasoners for RC. This scenario does not require any samples or training. We test the out-of-the-box performance of the PLM by directly prompting it with $\bm x_{prompt}$. Zero-shot in-context learning does not specify further hyperparameters since it is training-free.

\textbf{Zero-shot cross-lingual transfer}~~~In this scenario, following \citet{krishnan_2021_multilingual}, we fine-tune the model with in-language prompting on the English train split, and then conduct zero-shot in-context tests with this fine-tuned model on other languages using code-switch prompting. Through this setting, we want to verify if task-specific pre-training in a high-resource language such as English helps in other languages. In zero-shot cross-lingual transfer, we use the same hyperparameters and random seed to fine-tune on the English task.

\section{Results and Discussion}
\label{sec:results}
We first present the results in fully supervised, few-shot and zero-shot scenarios, and then discuss the main findings for answering the research questions in Section~\ref{sec:introduction}.

\insertfullysupervised

\subsection{Fully Supervised Results}
Table~\ref{tab:fully-supervised} presents the experimental results in the fully supervised scenario, for different methods, languages, and language groups. We see that all the three variants of our method beat the fine-tuning baseline \textsc{XLM-R}\textsubscript{\textsc{EM}} and the prompting baseline null prompts, according to the macro-averaged performance across 14 languages. In-language prompting delivers the most promising result, achieving an average $F_1$ of $85.0$, which is higher than \textsc{XLM-R}\textsubscript{\textsc{EM}} (68.2) and null prompts (66.2). The other two variants, code-switch prompting with and w/o soft tokens, achieve $F_1$ scores of 84.1 and 82.7, respectively, only 0.9 and 2.3 lower than in-language. All three prompt variants are hence effective in fully supervised scenarios.

On a per-group basis, we find that the lower-resourced a language is, the greater an advantage prompting enjoys against fine-tuning. In particular, in-language prompts shows better robustness compared to \textsc{XLM-R}\textsubscript{\textsc{EM}} in low-resource languages. They both yield 95.9-96.0 $F_1$ scores for English, but \textsc{XLM-R}\textsubscript{\textsc{EM}} decreases to 54.3 and 3.7 $F_1$ in Group-M and -L, while in-language prompting still delivers 83.5 and 65.2 $F_1$.

\subsection{Few-shot Results}
Table~\ref{tab:group} presents the per-group results in few-shot experiments. All the methods benefit from larger $K$. Similarly, in-language prompting still turns out to be the best contender, performing 1st in 8- and 32-shot, and the 2nd in 16-shot. We see that in-language outperforms \textsc{XLM-R}\textsubscript{\textsc{EM}} in all $K$-shots, while code-switch achieves comparable or even lower $F_1$ to \textsc{XLM-R}\textsubscript{\textsc{EM}} for $K=8$, suggesting that the choice of prompt affects the few-shot performance greatly, thus needs careful consideration.

On a per-group basis, we find that in-language prompting outperforms other methods for middle- and low-resourced languages. Similar observations can also be drawn from fully supervised results. We conclude that, with sufficient supervision, in-language is the optimal variant to prompt rather than code-switch. We hypothesize it is due to the pre-training scenario, where the PLM rarely sees code-mixed text \cite{santy_2021_bertologicomix}.
\insertfewshotgroup

\insertzeroshot

\subsection{Zero-shot Results}
Table~\ref{tab:zero-shot} presents the per-language results in zero-shot scenarios. We consider the random baseline for comparison~\cite{zhao_2021_discrete,winata_2021_language}. We notice that performance of the random baseline varies a lot across languages, since the languages have different number of classes in the dataset (cf.\ Table~\ref{tab:smiler}), with English being the hardest task.

For zero-shot in-context, code-switch prompting always outperforms the random baseline by a large margin, in both word orders, while in-language prompting performs worse than the random baseline in 6 languages. Code-switch prompting outperforms in-language prompting across all the 13 non-English languages, using SVO-template. We assume that, without in-language training, the PLM understands the task best when prompted in English. The impressive performance of code-switch shows the PLM is able to transfer its pre-trained knowledge in English to other languages. We also find that the performance is also highly indicated by the number of classes, with worst $F_1$ scores achieved in EN, KO and PT (36, 28 and 22 classes), and best scores in AR, RU and UK (9, 8 and 7 classes). In addition, we observe that word order does not play a significant role for most languages, except for FA, which is an SOV-language and has 54.5 $F_1$ gain from in-language prompting with an SOV-template.

For zero-shot cross-lingual transfer, we see that non-English tasks benefit from English in-domain prompt-based fine-tuning, and the $F_1$ gain improves with the English data size. For 5 languages (ES, FA, NL, SV, and UK), zero-shot transfer after training on 268k English examples delivers even better results than in-language fully supervised training (cf.\ Table~\ref{tab:fully-supervised}). \citet{sanh_2022_t0} show that including RC-specific prompt input in English during pre-training can help in other languages.

\subsection{Discussion}
Based on the results above, we answer the research questions from Section~\ref{sec:introduction}.

\textbf{RQ1.} Which is the most effective way to prompt? 
In the fully-supervised and few-shot scenario, in-language prompting displays the best results.
This appears to stem from a solid performance across all languages in both settings. Its worst performance is 31.8 $F_1$ for Polish 8-shot (see Table \ref{tab:few-shot-variance} in Appendix \ref{sec:additional_tables}).
All other methods have results lower than 15.0 $F_1$ for some language.
This indicates that with little supervision mT5 is able to perform the task when prompted in the language of the original text.
However, zero-shot results strongly prefer code-switch prompting.
It could follow that, without fine-tuning, the model's understanding of this task is much better in English.

\textbf{RQ2.} How well does our method perform in different data regimes and languages? 
Averaged over all languages, all our variants outperform the baselines, except for 8-shot.
For some high-resource languages, \textsc{XLM-R}\textsubscript{\textsc{EM}} is able to outperform our method.
On the other hand, for low-resource languages null prompts are a better baseline which we consistently outperform.
This could indicate that prompting the underlying mT5 model is better suited for multilingual RC on SMiLER.
Overall, the results suggest that minimal translation can be very helpful for multilingual relation classification.

\section{Related Work}
\label{sec:related-work}
\textbf{Multilingual relation classification}~~~Previous work in multilingual RC has primarily focused on traditional methods rather than prompting PLMs. \citet{faruqui-kumar_2015_multilingual} machine-translate non-English full text to English to deal with multilinguality.
\citet{akbik_2016_multilingual} employ a shared semantic role labeler to get language-agnostic abstraction and apply rule-based methods to classify the unified abstractions.
\citet{lin_2017_neural} employ convolutional networks to extract relation embeddings from texts, and propose cross-lingual attention between relation embeddings to model cross-lingual information consistency.
\citet{sanh_2019_hierarchical} leverage the embeddings from BiLSTM, which is trained with a set of selected semantic tasks to help (multilingual) relation extraction.
\citet{koksal-ozgur_2020_relx} fine-tune (multilingual) BERT, classifying the embedding at \texttt{[CLS]}. To take entity-related embeddings into consideration as well, \citet{nag_2021_data} add an extra summarization layer on top of a multilingual BERT to collect and pool the embeddings at both \texttt{[CLS]} and entity starts. 

\noindent\textbf{Multilingual prompting}~~~Multilingual prompting is a new yet fast-growing topic.
\citet{winata_2021_language} reduce handcrafting efforts by reformulating general classification tasks into binary classification with answers restricted to true or false for all languages. \citet{huang_2022_zero} propose a unified multilingual prompt by introducing a so-called ``two-tower'' encoder, with the template tower producing language-agnostic prompt representation, and the context tower encoding text information. \citet{fu_2022_polyglot} manually translate prompts and suggest multilingual multitask training to boost the performance for a target downstream task.

\section{Conclusion}
In this paper, we present a first, simple yet efficient and effective prompt method for multilingual relation classification, by translating only the relation labels.
Our prompting outperforms fine-tuning and null prompts in fully supervised and few-shot experiments.
With supervised data, in-language prompting enjoys the best performance, while in the zero-shot scenarios prompting in English is preferable. 
We attribute the good performance of our method to its well-suitedness for RC, with the derivation of \emph{entity\textsubscript{1}-\underline{relation}-entity\textsubscript{2}} prompts from relation triples. 
We would like to see our method extended to similar tasks, such as semantic role labeling, with a structure between concepts that can be described in natural language.

\section*{Limitations}
We acknowledge the main limitation of this work is that we only experiment on one dataset with 14 languages.
Multilingual RC datasets prior to SMiLER are limited in the coverage of languages or in the size of unique training examples.
It would be interesting to see how our method performs on other multilingual RC datasets, especially for underrepresented languages~\cite{winata_2022_nusax}.

We restrict the target language to be supported by the underlying PLM.
The popular multilingual PLMs, mT5 and mBART, include 101 and 25 languages during pre-training.
We rely on these PLMs and fail to study true low-resource languages that are not represented in such PLMs~\cite{aji_2022_one}.

It is noticeable that in the fully supervised scenario, for 7 out of the 14 languages, at least one method achieves over 0.95 micro-$F_1$ score.
We hypothesize that is due to high homogeneity in and between the train and test split.
If so, the dataset itself might not be challenging, which could indicate that the results are mostly measuring how well the model is able to fit a few indicators (quickly).
    
Like most other prompt methods, ours requires the label names to be natural language which are indicative of the class.
Therefore, our method would suffer from labels being non-descriptive.

\section*{Ethics Statement}
We use automated machine translation by Google Translate and DeepL for our method.
These MT systems contain biases regarding, e.g., gender (``has-author'': ``hat Autor'') where gender-neutral English nouns are translated to gendered nouns in target languages.

In this work we evaluate SMiLER~\cite{seganti_2021_smiler}, which is crawled from Wikipedia. In the paper, they have not stated measures that prevent collecting sensitive text. Therefore, we do not rule out the possible risk of sensitive content in the data.

The PLMs involved in this paper are BERT\textsubscript{\textsc{Base}} for EN(B), \textsc{XLM-R}\textsubscript{\textsc{Base}} for \textsc{XLM-R}\textsubscript{\textsc{EM}}, and mT5\textsubscript{\textsc{Base}} for null prompts and ours.
BERT\textsubscript{\textsc{Base}} is pre-trained on the BooksCorpus \citep{zhu2015aligning} and English Wikipedia.
\textsc{XLM-R}\textsubscript{\textsc{EM}} is pre-trained on a CommonCrawl corpus.
mT5\textsubscript{\textsc{Base}} is pre-trained on mC4, a filtered CommonCrawl corpus.
All our published models may have inherited biases from these corpora. 

\section*{Acknowledgments}
We would like to thank Nils Feldhus and the anonymous reviewers for their valuable comments and feedback on the paper. This work has been supported by the German Federal Ministry for Economic Affairs and Climate Action as part of the project PLASS (01MD19003E), and by the German Federal Ministry of Education and Research as part of the projects CORA4NLP (01IW20010) and BBDC2 (01IS18025E).

\bibliography{references}
\bibliographystyle{acl_natbib}

\appendix
\section{Experimental Details}
\label{sec:experimental-details}

\subsection{Hyperparameter Search}
We investigated the following possible hyperparameters for few-shot settings.
For fully-supervised, we take hyperparameters from literature (see Section \ref{sec:fully-supervised-setup}).

Number of epochs: $[10, 20]$; Learning rate: $[1\times10^{-5}, 3\times10^{-5}, 1\times10^{-4}, 3\times10^{-4}]$. Batch size: $[16, 64, 256]$, not tuned but selected based on available GPU VRAM.
\vspace{0.5em}

We manually tune these hyperparameters, based on the micro-$F_1$ score on the validation set.

\subsection{Computing Infrastructure}
Fully supervised experiments are conducted on a single A100-80GB GPU. Few-shot and zero-shot experiments are conducted on a single A100 GPU.

\subsection{Average Running Time}
\textbf{Fully supervised}~~~It takes 5 hours to train for 1 run with mT5\textsubscript{\textsc{BASE}} and a prompt method (null prompts, CS, SP and IL) on either English, or all other languages in total. With \textsc{XLM-R}\textsubscript{\textsc{EM}} the running time is 3 hours.

\textbf{Few-shot}~~~It takes 20 (8-shot), 26 (16-shot), and 36 minutes (32-shot) for 1 run with mT5\textsubscript{\textsc{BASE}} and a prompt method over all languages. With \textsc{XLM-R}\textsubscript{\textsc{EM}} the running time is 8 minutes.

\textbf{Zero-shot}~~~For zero-shot in-context experiments, it takes 6 minutes with mT5\textsubscript{\textsc{BASE}} and a  prompt method over all languages. For zero-shot cross-lingual transfer, the running time equals English training time (5 hours) plus inference-only time (6 minutes).
\vspace{0.5em}

\section{Verbalizers for SMiLER}
\label{sec:verbalizers}
\begin{itemize}
    \item \textbf{EN}~~~
        "birth-place": "birth place",
        "eats": "eats",
        "event-year": "event year",
        "first-product": "first product",
        "from-country": "from country",
        "has-author": "has author",
        "has-child": "has child",
        "has-edu": "has education",
        "has-genre": "has genre",
        "has-height": "has height",
        "has-highest-mountain": "has highest mountain",
        "has-length": "has length",
        "has-lifespan": "has lifespan",
        "has-nationality": "has nationality",
        "has-occupation": "has occupation",
        "has-parent": "has parent",
        "has-population": "has population",
        "has-sibling": "has sibling",
        "has-spouse": "has spouse",
        "has-tourist-attraction": "has tourist attraction",
        "has-type": "has type",
        "has-weight": "has weight",
        "headquarters": "headquarters",
        "invented-by": "invented by",
        "invented-when": "invented when",
        "is-member-of": "is member of",
        "is-where": "located in",
        "loc-leader": "location leader",
        "movie-has-director": "movie has director",
        "no\_relation": "no relation",
        "org-has-founder": "organization has founder",
        "org-has-member": "organization has member",
        "org-leader": "organization leader",
        "post-code": "post code",
        "starring": "starring",
        "won-award": "won award";
    \vspace{-0.5cm}
    \begin{figure}[h]
    \centering  \includegraphics[width=0.48\textwidth]{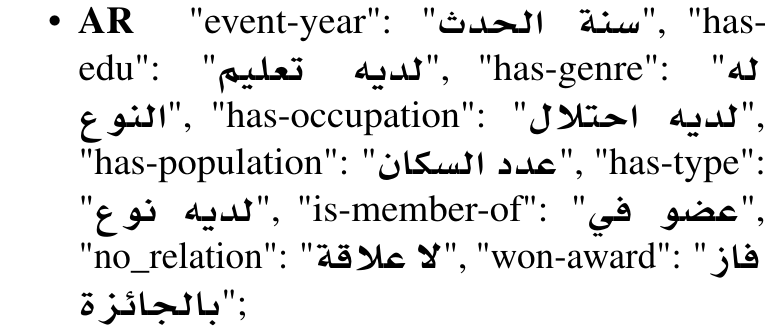}
    \label{fig:ar-verbalizers}
    \end{figure}
    \vspace{-1cm}

    \item \textbf{DE}~~~
    "birth-place": "Geburtsort",
    "event-year": "Veranstaltungsjahr",
    "from-country": "vom Land",
    "has-author": "hat Autor",
    "has-child": "hat Kind",
    "has-edu": "hat Bildung",
    "has-genre": "hat Genre",
    "has-occupation": "hat Beruf",
    "has-parent": "hat Elternteil",
    "has-population": "hat Bevölkerung",
    "has-spouse": "hat Ehepartner",
    "has-type": "hat Typ",
    "headquarters": "Hauptsitz",
    "is-member-of": "ist Mitglied von",
    "is-where": "gelegen in",
    "loc-leader": "Standortleiter",
    "movie-has-director": "Film hat Regisseur",
    "no\_relation": "keine Beziehung",
    "org-has-founder": "Organisation hat Gründer",
    "org-has-member": "Organisation hat Mitglied",
    "org-leader": "Organisationsleiter",
    "won-award": "gewann eine Auszeichnung";
    
    \item \textbf{ES}~~~
        "birth-place": "lugar de nacimiento",
        "event-year": "año del evento",
        "from-country": "del país",
        "has-author": "tiene autor",
        "has-child": "tiene hijo",
        "has-edu": "tiene educación",
        "has-genre": "tiene género",
        "has-occupation": "tiene ocupación",
        "has-parent": "tiene padre",
        "has-population": "tiene población",
        "has-spouse": "tiene cónyuge",
        "has-type": "tiene tipo",
        "headquarters": "sede central",
        "is-member-of": "es miembro de",
        "is-where": "situado en",
        "loc-leader": "líder de ubicación",
        "movie-has-director": "película cuenta con el director",
        "no\_relation": "sin relación",
        "org-has-founder": "organización cuenta con el fundador",
        "org-has-member": "organización tiene miembro",
        "won-award": "ganó el premio";
    \begin{figure}[h]
    \centering  \includegraphics[width=0.49\textwidth]{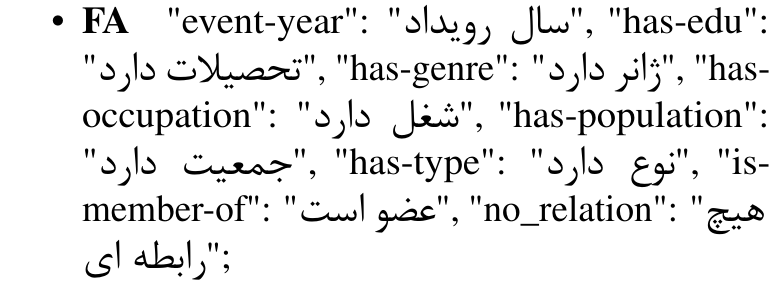}
    \label{fig:fa-verbalizers}
    \end{figure}
    \vspace{-1cm}

    \item \textbf{FR}~~~
        "birth-place": "lieu de naissance",
        "event-year": "année de l'événement",
        "from-country": "du pays",
        "has-author": "a un auteur",
        "has-child": "a un enfant",
        "has-edu": "a une éducation",
        "has-genre": "a un genre",
        "has-occupation": "a une profession",
        "has-parent": "a un parent",
        "has-population": "a de la population",
        "has-spouse": "a un conjoint",
        "has-type": "a le type",
        "headquarters": "siège social",
        "is-member-of": "est membre de",
        "is-where": "situé à",
        "loc-leader": "guide d'emplacement",
        "movie-has-director": "le film a un réalisateur",
        "no\_relation": "aucune relation",
        "org-has-founder": "l'organisation a un fondateur",
        "org-has-member": "l'organisation a un membre",
        "org-leader": "chef d'organisation",
        "won-award": "a remporté le prix";
    \item \textbf{IT}~~~
        "birth-place": "luogo di nascita",
        "event-year": "anno dell'evento",
        "from-country": "dal paese",
        "has-author": "ha autore",
        "has-child": "ha un figlio",
        "has-edu": "ha un'educazione",
        "has-genre": "ha genere",
        "has-occupation": "ha occupazione",
        "has-parent": "ha un genitore",
        "has-population": "ha una popolazione",
        "has-spouse": "ha un coniuge",
        "has-type": "ha il tipo",
        "headquarters": "sede centrale",
        "is-member-of": "è membro di",
        "is-where": "situato in",
        "loc-leader": "leader della posizione",
        "movie-has-director": "il film ha direttore",
        "no\_relation": "nessuna relazione",
        "org-has-founder": "l'organizzazione ha fondatore",
        "org-has-member": "l'organizzazione ha un membro",
        "org-leader": "leader dell'organizzazione",
        "won-award": "ha vinto un premio";
    \item \textbf{KO}~~~
        "birth-place": "출생지",
        "event-year": "이벤트 연도",
        "first-product": "첫 번째 제품",
        "from-country": "나라에서",
        "has-author": "저자가 있다",
        "has-child": "아이가 있다",
        "has-edu": "교육이 있다",
        "has-genre": "장르가 있다",
        "has-highest-mountain": "가장 높은 산이 있다",
        "has-nationality": "국적이 있다",
        "has-occupation": "직업이 있다",
        "has-parent": "부모가 있다",
        "has-population": "인구가 있다",
        "has-sibling": "형제가 있다",
        "has-spouse": "배우자가 있다",
        "has-tourist-attraction": "관광명소가 있다",
        "has-type": "유형이 있습니다",
        "headquarters": "본부",
        "invented-by": "에 의해 발명",
        "invented-when": "언제 발명",
        "is-member-of": "의 회원입니다",
        "is-where": "어디에",
        "movie-has-director": "영화에 감독이 있다",
        "no\_relation": "관계가 없다",
        "org-has-founder": "조직에는 설립자가 있습니다",
        "org-has-member": "조직에 구성원이 있습니다",
        "org-leader": "조직 리더",
        "won-award": "수상";
    \item \textbf{NL}~~~
    "birth-place": "geboorteplaats",
        "event-year": "evenementenjaar",
        "from-country": "van het land",
        "has-author": "heeft auteur",
        "has-child": "heeft kind",
        "has-edu": "heeft onderwijs",
        "has-genre": "heeft genre",
        "has-occupation": "heeft beroep",
        "has-parent": "heeft ouder",
        "has-population": "heeft bevolking",
        "has-spouse": "heeft echtgenoot",
        "has-type": "heeft type",
        "headquarters": "hoofdkantoor",
        "is-member-of": "is lid van",
        "is-where": "gevestigd in",
        "loc-leader": "locatieleider",
        "movie-has-director": "film had regisseur",
        "no\_relation": "geen relatie",
        "org-has-founder": "organisatie heeft oprichter",
        "org-has-member": "organisatie heeft lid",
        "org-leader": "organisatieleider",
        "won-award": "won prijs";
    \item \textbf{PL}~~~
        "birth-place": "miejsce urodzenia",
        "event-year": "rok imprezy",
        "from-country": "z kraju",
        "has-author": "ma autor",
        "has-child": "ma dziecko",
        "has-edu": "ma wykształcenie",
        "has-genre": "ma gatunek",
        "has-occupation": "ma zawód",
        "has-parent": "ma rodzica",
        "has-population": "ma ludność",
        "has-spouse": "ma współmałżonka",
        "has-type": "ma typ",
        "headquarters": "siedziba główna",
        "is-member-of": "jest członkiem",
        "is-where": "mieszczący się w",
        "loc-leader": "lider lokalizacji",
        "movie-has-director": "film ma reżysera",
        "org-has-founder": "organizacja ma założyciela",
        "org-has-member": "organizacja ma członków",
        "org-leader": "lider organizacji",
        "won-award": "otrzymał nagrodę";
    \item \textbf{PT}~~~
        "birth-place": "local de nascimento",
        "event-year": "ano do evento",
        "from-country": "do país",
        "has-author": "tem autor",
        "has-child": "tem filho",
        "has-edu": "tem educação",
        "has-genre": "tem género",
        "has-occupation": "tem ocupação",
        "has-parent": "tem pai",
        "has-population": "tem população",
        "has-spouse": "tem cônjuge",
        "has-type": "tem tipo",
        "headquarters": "sede",
        "is-member-of": "é membro de",
        "is-where": "localizado em",
        "loc-leader": "loc leader",
        "movie-has-director": "filme tem realizador",
        "no\_relation": "sem relação",
        "org-has-founder": "organização tem fundador",
        "org-has-member": "organização tem membro",
        "org-leader": "líder da organização",
        "won-award": "ganhou prémio";
    \item \textbf{RU}~~~
    "event-year": "\foreignlanguage{russian}{год события}",
    "has-edu": "\foreignlanguage{russian}{имеет образование}",
    "has-genre": "\foreignlanguage{russian}{имеет жанр}",
    "has-occupation": "\foreignlanguage{russian}{имеет профессию}",
    "has-population": "\foreignlanguage{russian}{имеет население}",
    "has-type": "\foreignlanguage{russian}{имеет тип}",
    "is-member-of": "\foreignlanguage{russian}{является членом}",
    "no\_relation": "\foreignlanguage{russian}{без связи}";
    \item \textbf{SV}~~~
        "birth-place": "födelseort",
        "event-year": "År för evenemanget",
        "from-country": "från ett land",
        "has-author": "har en författare",
        "has-child": "har chili",
        "has-edu": "har utbildning",
        "has-genre": "har en genre",
        "has-occupation": "har ockuperat",
        "has-parent": "har en förälder",
        "has-population": "har en befolkning",
        "has-spouse": "har make eller maka",
        "has-type": "har typ",
        "headquarters": "huvudkontor",
        "is-member-of": "är medlem i",
        "is-where": "som ligger i",
        "loc-leader": "platsansvarig",
        "movie-has-director": "filmen har regissör",
        "no\_relation": "ingen relation",
        "org-has-founder": "organisationen har en grundare",
        "org-has-member": "organisationen har en medlem",
        "org-leader": "ledare för organisationen",
        "won-award": "vann ett pris";
    \item \textbf{UK}~~~
    "event-year": "\foreignlanguage{ukrainian}{рік події}", 
    "has-edu": "\foreignlanguage{ukrainian}{має освіту}", 
    "has-genre": "\foreignlanguage{ukrainian}{має жанр}", 
    "has-occupation": "\foreignlanguage{ukrainian}{має заняття}", 
    "has-population": "\foreignlanguage{ukrainian}{має населення}", 
    "has-type": "\foreignlanguage{ukrainian}{має тип}", 
    "no\_relation": "\foreignlanguage{ukrainian}{ніякого відношення}".
\end{itemize}

\cleardoublepage
\section{Detailed Few-shot Results}
\label{sec:additional_tables}
\insertfewshotvariance

\end{document}